\documentclass[twoside]{article}

\usepackage[preprint]{aistats2026}
\usepackage[utf8]{inputenc}
\usepackage[T1]{fontenc}
\usepackage{amssymb}
\usepackage{amsmath,amsfonts,amsthm}
\usepackage{algorithm}
\usepackage{xspace}
\usepackage{algpseudocode}
\usepackage{xcolor}
\usepackage{graphicx}
\usepackage{booktabs}
\usepackage{siunitx}
\usepackage{adjustbox}
\usepackage{multirow}
\usepackage{enumitem}
\usepackage{subcaption}
\usepackage[acronym,nowarn]{glossaries}
\setacronymstyle{long-sc-short}
\glsdisablehyper
\makeglossaries

\usepackage[dvipsnames,table]{xcolor}

\usepackage{hyperref}
\hypersetup{colorlinks,citecolor=blue!50!green,linkcolor=red!80!black,urlcolor=blue}
\usepackage[capitalize,nameinlink]{cleveref}

\newcommand{\cL}{\mathcal{L}}

\newcommand{\cP}{\mathcal{P}}

\newcommand{\cR}{\mathcal{R}}

\newcommand{\cE}{\mathcal{E}}

\newcommand{\cX}{\mathcal{X}}
\newcommand{\cY}{\mathcal{Y}}
\newcommand{\cH}{\mathcal{H}}
\newcommand{\cZ}{\mathcal{Z}}
\newcommand{\cO}{\mathcal{O}}

\newcommand{\E}{\mathbb{E}}
\newcommand{\R}{\mathbb{R}}

\newcommand{\supp}{\textrm{supp}}

\newcommand{\KL}{\text{KL}}

\DeclareMathOperator{\Var}{Var}

\theoremstyle{plain}
\newtheorem{theorem}{Theorem}[section]
\newtheorem{lemma}[theorem]{Lemma}
\newtheorem{proposition}[theorem]{Proposition}

\theoremstyle{definition}
\newtheorem{assumption}[theorem]{Assumption}
\newtheorem{definition}[theorem]{Definition}
\theoremstyle{remark}
\newtheorem{remark}[theorem]{Remark}

\newacronym{IRM}{irm}{Invariant Risk Minimization}
\newacronym{MMD}{mmd}{Maximum Mean Discrepancy}
\newacronym{MLP}{mlp}{multi-layer perceptron}
\newacronym{SOTA}{sota}{state-of-the-art}
\newacronym{DA}{da}{Domain Adaptation}

\newcommand{\rex}{\textsc{re}{\small{x}}\xspace}
\newcommand{\irm}{\textsc{irm}\xspace}
\newcommand{\groupdro}{\textsc{g}{\small{roup}}{\textsc{dro}}\xspace}
\newcommand{\coral}{\textsc{coral}\xspace}
\newcommand{\uaed}{\textsc{uaed}\xspace}
\newcommand{\ibirm}{\textsc{ib}-\textsc{irm}\xspace}
\newcommand{\dfr}{\textsc{dfr}\xspace}
\newcommand{\disc}{\textsc{disc}\xspace}
\newcommand{\pair}{\textsc{pair}\xspace}
\newcommand{\jtt}{\textsc{jtt}\xspace}
\newcommand{\rvp}{\textsc{rvp}\xspace}
\newcommand{\mixup}{\textsc{mixup}\xspace}
\newcommand{\lisa}{\textsc{lisa}\xspace}
\newcommand{\eiil}{\textsc{eiil}\xspace}
\newcommand{\augerino}{\textsc{a}\textsc{ugerino}\xspace}
\newcommand{\autoaugment}{\textsc{autoaugment}\xspace}
\newcommand{\randaugment}{\textsc{randaugment}\xspace}
\newcommand{\augmax}{\textsc{augmax}\xspace}
\newcommand{\erm}{\textsc{erm}\xspace}
\newcommand{\birm}{\textsc{birm}\xspace}
\newcommand{\metairm}{\textsc{m}{\small{eta}}-\textsc{irm}\xspace}

\newcommand{\arex}{\textsc{a}-\textsc{re}{\small{x}}\xspace}
\newcommand{\airm}{\textsc{a}-\textsc{irm}\xspace}
\newcommand{\agroupdro}{\textsc{a}-\textsc{g}{\small{roup}}{\textsc{dro}}\xspace}
\newcommand{\acoral}{\textsc{a}-\textsc{coral}\xspace}

\newcommand{\resneteighteen}{\textsc{r}{\small{es}}\textsc{n}{\small{et18}}\xspace}
\newcommand{\resnetfifty}{\textsc{r}{\small{es}}\textsc{n}{\small{et50}}\xspace}
\newcommand{\waterbirds}{\textsc{w}{\small{ater}}\textsc{b}{\small{irds}}\xspace}
\newcommand{\imagenet}{\textsc{imagenet}\xspace}
\newcommand{\coloredmnist}{\textsc{colored}-\textsc{mnist}\xspace}
\newcommand{\rotatedmnist}{\textsc{rotated}-\textsc{mnist}\xspace}
\newcommand{\cmnist}{\textsc{c}-\textsc{mnist}\xspace}
\newcommand{\rmnist}{\textsc{r}-\textsc{mnist}\xspace}

\usepackage{natbib}
\begin{document}

%

%

\twocolumn[

\aistatstitle{Universal Adaptive Environment Discovery}

\renewcommand{\thefootnote}{\fnsymbol{footnote}}

\aistatsauthor{ Madi Matymov\footnotemark[1] \And Ba-Hien Tran \And  Maurizio Filippone }

\aistatsaddress{ KAUST \\ Saudi Arabia \And  Huawei Technologies France SASU \\ France \And KAUST \\ Saudi Arabia } ]

\renewcommand{\thefootnote}{\fnsymbol{footnote}}
\footnotetext[1]{Corresponding author: \texttt{madi.matymov@kaust.edu.sa}}

\begin{abstract}
An open problem in Machine Learning is how to avoid models to exploit spurious correlations in the data; a famous example is the background–label shortcut in the Waterbirds dataset. 
A common remedy is to train a model across multiple \emph{environments}; in the Waterbirds dataset, this corresponds to training by randomizing the background. 
However, selecting the right environments is a challenging problem, given that these are rarely known a priori. 
We propose \emph{Universal Adaptive Environment Discovery} (UAED), a unified framework that \emph{learns} a distribution over data transformations that instantiate environments, and optimizes any robust objective \emph{averaged} over this learned distribution. 
UAED yields adaptive variants of IRM, REx, GroupDRO, and CORAL without predefined groups or manual environment design. 
We provide a theoretical analysis by providing PAC-Bayes bounds and by showing robustness to test environment distributions under standard conditions.
Empirically, UAED discovers interpretable environment distributions and improves worst-case accuracy on standard benchmarks, while remaining competitive on mean accuracy. 
Our results indicate that making environments \emph{adaptive} is a practical route to out-of-distribution generalization.
\end{abstract}


\section{Introduction}

\begin{figure}[t]
\centering
\includegraphics[width=0.35\textwidth]{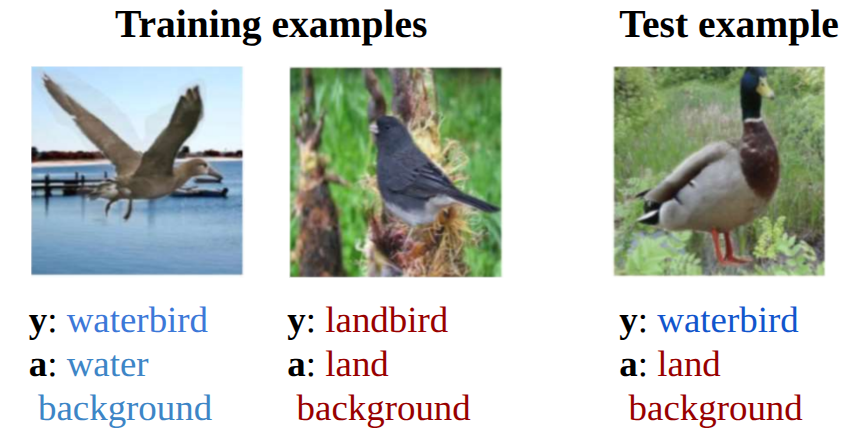}
\vspace{-1.5ex}
\caption{\small \waterbirds dataset examples \citep{sagawa2020waterbirds}: the training correlation between label $y$ and spurious attribute $a$ does not hold at test time.
}
\vspace{-1.75ex}
\label{fig:waterbirds_sample}

\end{figure}

Machine learning models often fail when deployed in environments that differ from their training conditions. For example, a medical diagnosis system trained on data from one hospital may perform poorly at another, and a model trained to recognize birds from professional photographs may struggle with amateur images. Such failures typically arise because models rely on \emph{spurious correlations} (see, e.g., \cite{sagawa2019distributionally}, and references therein)--patterns present in the training data that do not generalize to deployment settings.
\cref{fig:waterbirds_sample} illustrates this phenomenon on the \waterbirds dataset \citep{sagawa2020waterbirds}.
When a model learns spurious correlations, it can achieve high i.i.d. test accuracy but fail dramatically on subgroups where the correlation does not hold.

The machine learning community has developed numerous approaches to address this challenge. 
Invariant Risk Minimization 
\citep{arjovsky2019irm} seeks features with stable predictive relationships across environments.
Risk Extrapolation (\rex) \citep{krueger2021out} minimizes variance of risks across environments. Group Distributionally Robust Optimization (\groupdro) \citep{sagawa2019distributionally} optimizes worst-case performance over predefined groups. Domain alignment methods include covariance alignment via \coral, which matches second-order feature statistics across domains \citep{sun2016deep}, and kernel mean–matching approaches based on \gls{MMD} \citep{gretton12mmd}.


Despite their different objectives, all these methods share a critical limitation: they require practitioners to predefine environments or groups that appropriately expose spurious correlations. This creates a fundamental paradox—identifying which correlations are spurious requires knowing the environments, but defining good environments requires knowing which correlations are spurious.








\textbf{Our Contributions.} We introduce Universal Adaptive Environment Discovery (\uaed), a framework that makes environment specification learnable rather than fixed. Specifically, we: \textbf{(1)} \emph{ unify diverse robust methods} (\irm, \rex, \groupdro, \coral) under adaptive environment discovery, parameterized through learnable data transformations; \textbf{(2)} \emph{provide a theoretical foundation} via PAC-Bayes bounds showing distributionally robust guarantees with implicit regularization explaining method-specific behaviors; \textbf{(3)} \emph{validate empirically} on synthetic and real-world benchmarks, where adaptive variants consistently improve over baselines and approach state-of-the-art without predefined groups; and \textbf{(4)} \emph{offer the conceptual insight} that robust objectives and environment discovery are complementary, with the latter being critical to the former’s success.

Our work reveals that the dichotomy between ``choosing robust objectives'' and ``defining environments'' is artificial. By making environments learnable, we provide a more principled and practical path toward out-of-distribution generalization.

\section{Related Work}

\textbf{Robust Learning Methods.} There exist several approaches for robust learning, each with distinct objectives. \irm \citep{arjovsky2019irm} seeks invariant predictors but is highly sensitive to environment specification \citep{gulrajani2021search, rosenfeld2020risks}. Variants such as \ibirm \citep{ahuja2021invariance} and \pair \citep{kim2021selfieboosting} still require predefined environments. \rex \citep{krueger2021out} and Risk Variance Penalization (\rvp) \citep{xie2020risk} minimize variance across environments, while \groupdro \citep{sagawa2019distributionally} optimizes worst-case group performance. More recent methods, including Just Train Twice (\jtt) \citep{liu2021just}, \disc \citep{wu2023discovering}, and Deep Feature Reweighting (\dfr) \citep{kirichenko2023last}, achieve strong performance but rely on group annotations or specialized architectures.

\textbf{Environment Discovery.} Prior work has approached environment discovery from different perspectives. \eiil \citep{creager2021environment} leverages causal discovery but requires access to the underlying causal graph. Environment inference methods such as heterogeneous risk minimization \citep{liu2021heterogeneous} generally assume environments are discrete. \lisa \citep{yao2022improving} instead employs \mixup \citep{zhang2018mixup} for implicit environment augmentation. In contrast, our framework unifies these directions by enabling any robust objective to operate with learned \emph{continuous} environment distributions.

\textbf{Data Augmentation Learning.} \autoaugment \citep{cubuk2019autoaugment} and \randaugment \citep{cubuk2020randaugment} learn augmentation policies for standard training, while \augerino \citep{benton2020learning} learns invariances directly from data. \augmax \citep{wang2021augmax} employs adversarial augmentations to enhance robustness. Most relevant is targeted augmentation \citep{gao2023out}, which shows that augmentation can improve OOD performance. OPTIMA \citep{matymov2025optima} placed priors over transformation policies and performs evidence-driven selection to improve generalization under shift. We build on these insights by showing that augmentation learning and environment specification are fundamentally connected.

\textbf{Theoretical Foundations.} PAC-Bayes theory \citep{mcallester1999pac} provides generalization bounds that trade off empirical risk against model complexity. For domain adaptation, \citet{germain2013pac} derived bounds based on divergence measures. More recently, \citet{deng2021labels} established connections between PAC-Bayes and domain generalization. We extend this line of work by proving that adaptive environment discovery yields distributionally robust guarantees for any robust learning objective.

\section{Universal Adaptive Environment Discovery} \label{sec:unified_framework}

\paragraph{Setup.}
We consider inputs $X\in\cX$, labels/targets $Y\in\cY$, and environments $e\in\cE$.
Each environment induces a joint distribution $P^e$ on $\cX\times\cY$.
The goal is to learn a predictor $f_\theta:\cX\to\hat\cY$ that \emph{generalizes across environments}, even when the marginals $P^e(X)$ and spurious dependencies in $P^e(Y\mid X)$ vary with $e$.
We assume the existence of an underlying stable mechanism, formalized next.

\begin{assumption}[Invariant conditional]\label{ass:invariant}
There exists a representation $\Phi^*:\cX\to\cZ$ such that the conditional distribution of $Y$ given the representation is invariant across environments:
\[
P^e\!\big(Y \mid \Phi^*(X)\big) \;=\; P\!\big(Y \mid \Phi^*(X)\big)\qquad \text{for all } e\in\cE.
\]
\end{assumption}

\noindent\textbf{Remarks.}
(i) This is the standard invariant-prediction premise (e.g., \citep{peters2016causal,arjovsky2019irm}); it is \emph{task-agnostic} and covers classification and regression (no specific link function is assumed).
(ii) We use it to motivate environment diversity; our PAC--Bayes and DRO guarantees do \emph{not} rely on Assumption~\ref{ass:invariant}.

\subsection{Baseline robust objectives (fixed environments)}
Let $\cE_{\text{train}}=\{e_1,\dots,e_k\}$ be given. For a bounded or sub-gamma loss $\ell$,
\[
\cR^e(\theta)\;=\;\E_{(x,y)\sim P^e}\,\ell\big(f_\theta(x),y\big).
\]
We write all baselines (except GroupDRO) as
\begin{align}
\label{eq:fixed-template}
\min_{\theta}\;
\underbrace{\frac{1}{k}\sum_{e\in\cE_{\text{train}}}\cR^e(\theta)}_{\text{mean risk}}
\;+\;
\eta\;\underbrace{\cP_{\text{robust}}^{\text{fixed}}(\theta;\cE_{\text{train}})}_{\text{method-specific regularizer}}.
\end{align}
\textbf{IRM (v1)} \citep{arjovsky2019irm}:
$$
\displaystyle
\cP_{\text{robust}}^{\text{fixed}}
= \frac{1}{k}\sum_{e}\left\|\,\nabla_{w}\,\cR^e\!\big(w\!\cdot\! f_\theta\big)\big|_{w=1}\right\|_2^2.
$$

\smallskip
\textbf{REx / VREx} \citep{krueger2021out}:
$$
\displaystyle
\cP_{\text{robust}}^{\text{fixed}}
= \Var_{e}\!\big[\cR^e(\theta)\big].
$$

\smallskip
\textbf{CORAL} \citep{sun2016deep}:
Let $F_\theta(\cdot)$ denote features. Using minibatch covariances,
{\small
$$
\displaystyle
\cP_{\text{robust}}^{\text{fixed}}
= \frac{1}{k(k-1)}\sum_{e\neq e'}
\left\|\,\mathrm{Cov}\big(F_\theta(X^e)\big)-\mathrm{Cov}\big(F_\theta(X^{e'})\big)\right\|_F^2.
$$
}



\smallskip
\textbf{GroupDRO} \citep{sagawa2019distributionally} (separate form):
$$
\displaystyle
\min_{\theta}\;\max_{e\in\cE_{\text{train}}}\;\cR^e(\theta),
$$
often optimized via the entropic surrogate
$
\displaystyle
\min_{\theta}\; \frac{1}{\lambda}\log\!\Big(\frac{1}{k}\sum_e e^{\lambda \cR^e(\theta)}\Big)$ with $\lambda>0$.

All these approaches assume access to predefined environments $\{e_1, \dots, e_k\}$, leading to the central challenge: \emph{how can we define environments without knowing a priori what distinguishes them?}

\paragraph{Policy over environments.}
Instead of fixing $\{e_1,\dots,e_k\}$, we index environments by a parameter $\gamma\in\Gamma$ (e.g., correlation strength, rotation angle, style).
A \emph{policy} $\Pi_\phi=p(\gamma\mid\phi)$ (density or categorical mass) over $\Gamma$ selects environments during training; let $e(\gamma)$ be the corresponding environment.


\subsection{The Unified UAED Framework}
Inspired by variational principles \citep{jordan1999variational}, we replace fixed environments with a learned distribution over environments. We define the policy-averaged risk as
\[
\cR_{\Pi_\phi}(\theta)\;=\;\E_{\gamma\sim\Pi_\phi}\;\E_{(x,y)\sim P}\;\ell\big(f_\theta(T_\gamma(x)),y\big),
\]
where $T_\gamma$ denotes the (possibly stochastic) data transformation that instantiates $e(\gamma)$.

\begin{definition}[Universal Adaptive Objective]\label{def:uaed}
Given a robust method with regularizer $\cP_{\text{robust}}$, UAED optimizes
\begin{align}
\label{eq:uaed-master}
\min_{\theta,\phi}\;
\textcolor{Blue}{\underbrace{\E_{\gamma\sim\Pi_\phi}\big[\cR^{\,e(\gamma)}(\theta)\big]}_{\text{policy-averaged risk}}}
\;+\;
\textcolor{Bittersweet}{\eta\,\underbrace{\cP_{\text{robust}}(\theta;\Pi_\phi)}_{\text{method-specific}}} \nonumber
\;+\;\\
+\textcolor{Plum}{\beta\,\KL\!\big(\Pi_\phi\big\Vert\Pi_0\big)},
\end{align}
where $\Pi_0$ is a fixed prior on $\Gamma$, and $\eta,\beta\!\ge\!0$.
\end{definition}

For each method, $\cP_{\text{robust}}(\theta;\Pi_\phi)$ is

\textbf{Adaptive IRM (A-IRM):}
$$
\displaystyle
\textcolor{Bittersweet}{\E_{\gamma\sim \Pi_\phi}\!\left\|\,\nabla_{w}\,\cR^{e(\gamma)}\!\big(w\!\cdot\! f_{\theta}\big)\big|_{w=1}\right\|_2^2.}
$$

\textbf{Adaptive REx (A-REx):}
$$
\displaystyle
\textcolor{Bittersweet}{\Var_{\gamma \sim \Pi_\phi}\!\big[\cR^{e(\gamma)}(\theta)\big].}
$$

\textbf{Adaptive CORAL (A-CORAL):}
{\small
$$
\textcolor{Bittersweet}{\E_{\gamma_1,\gamma_2\sim \Pi_{\phi}}
\left\|\,\mathrm{Cov}\big(F_\theta(X^{e(\gamma_1)})\big)-\mathrm{Cov}\big(F_\theta(X^{e(\gamma_2)})\big)\right\|_F^2.}
$$
}


\textbf{Adaptive GroupDRO (A-GroupDRO):}
Directly maximizing over $\gamma$ is non-smooth. Using the entropic-risk dual of a KL-ball DRO (Prop.~\ref{prop:entropic-duality-main}), we obtain a smooth surrogate:
\begin{align*}
\min_{\theta,\phi}\;
\textcolor{Blue}{\underbrace{\frac{1}{\lambda} \log
\mathbb{E}_{\gamma\sim \Pi_{\phi}}\exp\!\big\{\lambda\,\cR^{e(\gamma)}(\theta)\big\}}_{\text{entropic (smooth) worst-case}}}
\;+\;
\textcolor{Plum}{\beta\,\KL\!\big(\Pi_{\phi}\Vert\Pi_0\big)},
\end{align*}
for $\lambda>0$, optionally adding $\rho/\lambda$ to enforce robustness to a KL-ball of radius $\rho$ around $\Pi_\phi$.

\paragraph{Variational note.} When $\ell$ is a negative log-likelihood, the UAED objective in \eqref{eq:uaed-master} is (up to the method-specific regularizer and a temperature $\beta$ \citep{higgins2017betaVAE,bissiri2016general}) exactly a negative ELBO with prior $\Pi_0$ over $\gamma$ and variational posterior $\Pi_\phi$; our Loss Averaging (LoA)-Log-Likelihood Averaging (LA) \cref{prop:loa-la-main} bound justifies using the policy-averaged risk in place of the log-mixture.

\subsection{Implementation: Hierarchical Bayesian Framework}

We adopt a hierarchical Bayesian framework \citep[see, e.g.,][]{gelman1995bayesian} for the environment policy $p(\gamma \mid \phi)$ to encourage broad exploration and mitigate the risk of overfitting to specific, narrow environment.

\textbf{Continuous Environments:} For continuous $\gamma \in [0,1]$ (e.g., correlation strength), 
we model the policy as a Beta distribution whose shape and rate parameters are generated by a \gls{MLP}:
\begin{align}
\gamma &\sim \mathrm{Beta}(\alpha(\phi), \beta(\phi)),\\
\alpha(\phi), \beta(\phi) &= \mathrm{softplus}(\mathrm{MLP}(\phi)) + \varepsilon,
\end{align}
where $\varepsilon=10^{-6}$ is a small \emph{deterministic} offset used only for numerical stability.

\textbf{Discrete Environments:} For discrete transformations, the environment is modeled by a Categorical distribution with logits paramterized by $\phi$:
\begin{align}
p(\gamma|\phi) &= \text{Categorical}(\text{softmax}(\phi/\tau)).
\end{align}
To enable end-to-end backpropagation, we employ Gumbel-Softmax reparameterization \citep{jang2017categorical} to sample from this discrete distribution.
We employ a temperature schedule $\tau_t$ annealed linearly from $1.0$ to $0.3$ over the first $T_0$ epochs.


\subsection{Why Different Methods Need Different Environments}



Each robust objective implicitly seeks different environmental properties: \airm discovers environments with conflicting spurious correlations, \arex favors maximally diverse yet learnable environments, \agroupdro identifies stress-test scenarios, and \acoral aligns environments with distinct marginals. This flexibility to tailor the discovery process is the core advantage of our approach, enabling each method to automatically generate the environments needed for its objective to converge to the true invariant mechanism.

\vspace{-1.5ex}

\section{Theoretical Analysis} \label{sec:theoretical_analysis}

In this section, we give guarantees for \uaed that (i) are PAC--Bayes for the loss averaged under the \emph{learned} environment policy $\Pi_\phi$, and (ii) imply \emph{distributional robustness in environment space}: any test environment distribution inside a $\KL$--ball around $\Pi_\phi$ is controlled.

\paragraph{Setup.}
Let $X\in\cX$, $Y\in\cY$, $Z=(X,Y)\in\cZ=\cX\times\cY$, and $P$ a base distribution on $\cZ$.
Let $\Gamma$ index data transformations and $\Pi_\phi$ be a policy over $\Gamma$.
Each $\gamma\in\Gamma$ specifies a (possibly stochastic) map $T_\gamma$ that maps a point $z\in\cZ$ to a distribution over $z'\in\cZ$.
The induced environment is the \emph{pushforward} of $P$ by $T_\gamma$:
\begin{align}
& \hspace{5ex}P^{e(\gamma)} \;:=\; (T_\gamma)_\# P,
\qquad\text{meaning}\\&
\int f(z')\,dP^{e(\gamma)}(z')
= \int \Big(\E_{z'\sim T_\gamma(\cdot\mid z)} f(z')\Big)\, dP(z) \nonumber
\end{align}
for all bounded measurable $f$.
For a bounded loss $\ell:\cH\times\cZ\to[0,1]$ and predictor $h\in\cH$,
\begin{align}
\cR^{e(\gamma)}(h) &= \E_{z\sim P}\E_{z'\sim T_\gamma(\cdot\mid z)} \,\ell(h,z'),\\
\cR_{\Pi_\phi}(h) &= \E_{\gamma\sim\Pi_\phi}\,\cR^{e(\gamma)}(h).
\end{align}
Given $S=(z_1,\dots,z_n)\sim P^{\otimes n}$, the empirical risk is
\begin{align}
\hat\cR_{\Pi_\phi}(h)=\frac1n\sum_{i=1}^n \E_{\gamma\sim\Pi_\phi}\E_{z'\sim T_\gamma(\cdot\mid z_i)}\ell(h,z').
\end{align}

\noindent\textit{Readability note.} One can read $(T_\gamma)_\#P$ as “the distribution of the transformed sample $Z'$, obtained by applying $\gamma$ to $Z\sim P$.”

\paragraph{Joint prior and posterior.}
Treat the \emph{model} and the \emph{policy} as a single hypothesis $H=(h,\phi)$ with prior $M=P\times \Pi_0$ that is independent of $S$. Let $Q$ be any posterior over $(h,\phi)$ learned from $S$. All expectations $\E_{H\sim Q}[\cdot]$ are w.r.t.\ this joint posterior.

\begin{assumption}[Bounded or sub-gamma loss]\label{ass:bounded}
Either (a) $\ell\in[0,1]$, or (b) $\ell$ is sub-gamma under all $P^{e(\gamma)}$ with variance proxy $\sigma^2$ and scale $c>0$ (Remark~\ref{rem:subgamma-main}).
\end{assumption}

\subsection{PAC--Bayes under a learned policy}
\begin{theorem}[PAC--Bayes for policy-averaged risk]\label{thm:pacbayes-policy-main}
Under Assumption~\ref{ass:bounded} with bounded loss, with probability at least $1-\delta$ over $S$, for all posteriors $Q$,
\begin{align}
\textcolor{black}{\E_{H\sim Q}\big[\cR_{\Pi_\phi}(h)\big]}
\;\le\; &
\textcolor{black}{\E_{H\sim Q}\big[\hat\cR_{\Pi_\phi}(h)\big]}
+ \\
&+ \textcolor{black}{\sqrt{\frac{\KL(Q\Vert M)+\ln(1/\delta)}{2n}}}. \nonumber
\end{align}
\end{theorem}

\vspace{-1ex}

\noindent The theorem establishes a direct generalization guarantee for the $\Pi_\phi$--averaged risk by applying the standard PAC--Bayes bound \citep{mcallester1999pac} to the composite loss $(h,\phi,z)\mapsto \E_{\gamma\sim\Pi_\phi}\E_{z'|z}\ell(h,z')\in[0,1]$.
Since the policy $\phi$ is learned simultaneously with the model $h$, the \textcolor{black}{policy-averaged empirical risk $\E_{H\sim Q}[\hat\cR_{\Pi_\phi}(h)]$} serves as a sharp proxy for the \textcolor{black}{true policy-averaged risk $\E_{H\sim Q}[\cR_{\Pi_\phi}(h)]$}, controlled by the \textcolor{black}{complexity term $\KL(Q\Vert M)$}. 

\vspace{-1ex}

\subsection{Robustness to environment shift via KL--balls}
We now relate risks under any \emph{test} environment distribution $G$ over $\gamma$ to those under $\Pi_\phi$.

\begin{lemma}[Change of environment via DV+Hoeffding]\label{lem:transport-main}
Let $f:\Gamma\to[0,1]$ be measurable. For any $G$ and any policy $\Pi_\phi$,
\vspace{-1ex}
\[
\E_{\gamma\sim G} f(\gamma) \;\le\; \E_{\gamma\sim \Pi_\phi} f(\gamma) + \sqrt{\tfrac12\,\KL\!\big(G\Vert \Pi_\phi\big)}.
\]
\end{lemma}


\begin{theorem}[\uaed robust generalization]\label{thm:universal-robust-main}
Under Assumption~\ref{ass:bounded} with bounded loss, with probability at least $1-\delta$ over $S$, for all posteriors $Q$ and all $G$ satisfying $\KL(G\Vert \Pi_\phi)\le \rho$,
\begin{align}
\E_{H\sim Q}\E_{\gamma\sim G}\big[\cR^{e(\gamma)}(h)\big]
\;\le\;
\E_{H\sim Q}\big[\hat\cR_{\Pi_\phi}(h)\big] 
+\\
+\sqrt{\frac{\KL(Q\Vert M)+\ln(1/\delta)}{2n}}
+\sqrt{\frac{\rho}{2}}.  \nonumber
\end{align}
\end{theorem}

\paragraph{Interpretation.}
Minimizing the empirical \emph{policy-averaged} risk controls the risk under \emph{any} test environment mixture $G$ in a $\KL$--ball of radius $\rho$ around the learned policy $\Pi_\phi$, independently of which added robust penalty (\irm/\rex/\groupdro/\coral)  during training.

\subsection{Entropic risk duality (GroupDRO view)}
\begin{proposition}[KL--ball DRO equals entropic risk]\label{prop:entropic-duality-main}
Fix $h$ and write $r_\gamma=\cR^{e(\gamma)}(h)$. For any $\rho>0$,
\[
\sup_{G:\,\KL(G\Vert \Pi_\phi)\le \rho}\;\E_G[r_\gamma]
\;=\;
\inf_{\lambda>0}\;\frac{\rho + \log \E_{\gamma\sim\Pi_\phi}\exp\{\lambda r_\gamma\}}{\lambda}.
\]
Moreover, if $\Pi_\phi$ has finite support of size $k$ then, for any $\lambda>0$,
\vspace{-1ex}
\[
\max_{\gamma} r_\gamma
\;\le\;
\frac{1}{\lambda}\log\!\Big(\tfrac{1}{k}\sum_{\gamma}\! e^{\lambda r_\gamma}\Big) + \frac{\log k}{\lambda}.
\]
\end{proposition}

\vspace{-2.5ex}

\noindent Hence \emph{max} risk is upper-bounded by a log-sum-exp (entropic) risk—\textbf{not} by ``mean\,$+$\,\,$\sqrt{\Var}$'' in general.

\subsection{Loss-vs-likelihood averaging (safe replacement)}

The optimization objective of our \uaed framework is the minimization of the policy-averaged risk, $\cR_{\Pi_{\phi}}(h)$.
This risk corresponds to the Loss Averaging (LoA) objective.
In this section, we analyze the relationship between this quantity and Log-Likelihood Averaging (LA).
This comparison demonstrates that the LA quantity serves as a safe, variance controlled proxy for the LoA objective.

\begin{proposition}[LoA vs.\ LA gap is variance-controlled]\label{prop:loa-la-main}
Let $\ell_\gamma=\E_{z'|z}\ell(h,z')$ for fixed $(h,z)$ and denote $\mu=\E_{\Pi_\phi}\ell_\gamma$. We define:
\[
\cL_{\mathrm{LoA}}=\E_{\Pi_\phi}\ell_\gamma,\qquad
\cL_{\mathrm{LA}}=-\log \E_{\Pi_\phi}e^{-\ell_\gamma}.
\]
If $\ell_\gamma-\mu$ is sub-Gaussian with proxy $\sigma^2$, then
\[
0\le \cL_{\mathrm{LoA}}-\cL_{\mathrm{LA}}
=\log \E_{\Pi_\phi} e^{-(\ell_\gamma-\mu)} \le \sigma^2/2.
\]
\end{proposition}

\vspace{-2.5ex}

\paragraph{Interpretation.}
The proposition demonstrates a key theoretical advantage: minimizing $\cL_{\mathrm{LA}}$ is a safe replacement for minimizing $\cL_{\mathrm{LoA}}$, as the two are equivalent up to a term proportional to the loss variance $\sigma^2$.
This bounds mathematically confirms that optimizing an objective that is close to $\cL_{\mathrm{LoA}}$ (the policy-averaged risk) implicitly introduces a regularization against cross-environment loss variance. We will further provide empirical evidence in the experiments.

While the preceding theoretical analyses rely on the loss $l$ being bounded or sub-Gaussian, we can generalize these results to the sub-gamma condition covering a broader class of potentially heavy-tailed losses.

\begin{remark}[Sub-gamma variant]\label{rem:subgamma-main}
If $\ell$ is sub-gamma with variance proxy $\sigma^2$ and scale $c$, then in Theorem~\ref{thm:universal-robust-main} the PAC--Bayes term becomes its standard sub-gamma analogue, and the robustness term in Lemma~\ref{lem:transport-main} is replaced by $\inf_{\lambda\in(0,1/c)}\{\rho/\lambda+\psi(\lambda)\}$ with $\psi(\lambda)\le \frac{\sigma^2\lambda^2}{2(1-c\lambda)}$.
\end{remark}


\subsection{Optimization Strategy} \label{sec:optim_strategy}

Our \uaed framework requires minimizing a joint objective function that concurrently optimizes the model parameters $\theta$ (for hypothesis $h_{\theta}$ and the policy parameters $\phi$ (for environment distribution $\Pi_{\phi}$), as follows:
\begin{align}
L(\theta,\phi)=&\E_{\gamma\sim \Pi_\phi}\E_{z\sim P}\ell(h_\theta,T_\gamma(z)) +\\ &+ \eta\cP_{\text{robust}}(\theta;\Pi_\phi) + \beta  \,\KL(\Pi_\phi\Vert \Pi_0), \nonumber
\end{align}
where $\cP_{\text{robust}}$ represents the robust penalty (e.g., the \irm penalty).
Due to the coupled nature of $\theta$ and $\phi$ within the objective, we adopt an alternating optimization scheme. At each training step, we sample a mini-batch and update the policy parameters $\phi$ while holding $\theta$ fixed, and then update $\theta$ while keeping $\phi$ fixed.
\cref{app:alternating_optim} provides a convergence analysis of our optimization strategy.



\vspace{-1ex}

\section{Experiments} \label{sec:experiments}

\vspace{-1ex}

We evaluate our universal adaptive framework across both synthetic benchmarks and real-world datasets, demonstrating consistent improvements across diverse spurious correlation types.



\textbf{Datasets.} We use three standard benchmarks. \coloredmnist \citep{arjovsky2019irm} is a binary digit task ($<5$ vs.\ $\geq 5$) with spurious color correlations controlled by $e\in[0,1]$; models train on $e\in\{0.1,0.2\}$ and test on $e=0.9$. \rotatedmnist \citep{Ghifary_2015_ICCV} uses the same task with rotations $\{0^\circ,45^\circ,90^\circ,135^\circ,180^\circ\}$, training on $\{0^\circ,90^\circ\}$. \waterbirds~\citep{wah2011caltech,sagawa2020waterbirds} is constructed by compositing CUB-200-2011 foreground birds onto background scenes to induce spurious correlations between label and background; it defines four groups by (bird type, background).


\textbf{Implementation Details.} For the synthetic benchmarks, we use $3$-layer \glspl{MLP} with 256 hidden units, while for \waterbirds we employ \resneteighteen and \resnetfifty pretrained on \imagenet \citep{krizhevsky2012imagenet}. Models are optimized with an Adam optimizer \citep{kingma2014adam} with a learning rate of $10^{-4}$. The adaptive policy applies a learning rate multiplier of $100\times$ for continuous and $10\times$ for discrete settings. We sample environments using $5$ Monte Carlo samples for synthetic datasets and $3$ for \waterbirds, and report worst-case accuracy across environments or groups for evaluation.

\subsection{A-IRM Results on Synthetic Benchmarks}

\paragraph{A-IRM vs. standard IRM.}
\cref{tab:airm_synthetic} demonstrates that our \airm significantly outperforms standard \irm on both synthetic benchmarks, with remarkable improvements on \rotatedmnist (+28.4\%). The lower variance of \airm indicates more stable training.

\begin{table}[h]
\centering
\caption{\small{\airm results on \coloredmnist and \rotatedmnist.}}
\label{tab:airm_synthetic}
\vspace{-1.5ex}
\renewcommand{\arraystretch}{1.00}
\scalebox{.78}{
    \begin{tabular}{llcc}
    \toprule
    \textbf{Dataset} & \textbf{Method} & \textbf{Worst-Case Acc } & \textbf{Gain} \\
    \midrule
    \midrule
    \multirow{2}{*}{\cmnist} 
        & \irm ($e \in \{0.1, 0.2\}$) & $66.8 \pm 2.9$ (\%) & — \\
        & \airm & $\mathbf{72.3 \pm 1.6}$ (\%) & +5.5\% \\
    \midrule
    \multirow{2}{*}{\rmnist} 
        & \irm ($\{0°, 90°\}$) & $65.8 \pm 0.7$ (\%) & — \\
        & \airm & $\mathbf{94.2 \pm 0.2}$ (\%) & +28.4\% \\
    \bottomrule
    \end{tabular}
}
\vspace{-1.5ex}
\end{table}

\paragraph{Comparison with IRM variants.}
Moreover, \airm achieves competitive performance with specialized \irm variants and matches the performance of manually optimized oracle \irm setup (see \cref{tab:irm_variants}). Specifically, \airm achieves the highest accuracy on the challenging \coloredmnist test set ($e = 0.9$). Its performance matches the manually-tuned \irm ($e \in \{0.2, 0.8\}$), demonstrating that the adaptive policy automatically discovers optimal environments.


\begin{table}[h]
\centering
\caption{\small{Comparison with \irm variants on \coloredmnist (test $e = 0.9$). The gains are compared to \irm.}}
\vspace{-1.5ex}
\label{tab:irm_variants}
\renewcommand{\arraystretch}{1.00}
\scalebox{.88}{
\begin{tabular}{lcc}
\toprule
\textbf{Method} & \textbf{Test Acc (\%)} & \textbf{Gain} \\
\midrule
\midrule
\erm & $17.1 \pm 0.6$ & -74.4\% \\
\irm \citep{arjovsky2019irm} & $66.8 \pm 2.9$ & — \\
\metairm \citep{bae2021meta} & $70.5 \pm 0.9$ & +5.5\% \\
\birm \citep{lin2022bayesian} & $69.8 \pm 1.2$ & +4.5\% \\
\irm ($e \in \{0.2, 0.8\}$) & $72.2 \pm 0.5$ & +8.1\% \\
\airm \textbf{(Ours)} & $\mathbf{72.3 \pm 1.6}$ & \textbf{+8.2\%} \\
\bottomrule
\end{tabular}
}
\vspace{-2ex}
\end{table}


\subsection{Environment Discovery Analysis}

\begin{figure*}[t!]
\centering
\includegraphics[width=0.77\textwidth]{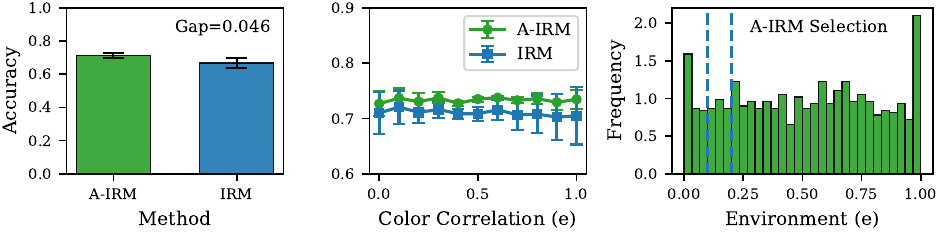}
\vspace{-1.5ex}
\caption{\small \airm environment discovery on \coloredmnist. \textbf{(Left)} Worst-case test accuracy—\airm achieves 72.3\% vs \irm's 66.8\%, a 5.5\% improvement. \textbf{(Middle)} Test accuracy across different color correlations shows \airm's flatter profile, indicating successful invariant learning. \textbf{(Right)} \airm discovers intermediate correlations ($e \approx 0.35$) rather than the extremes used by standard \irm (blue dashed lines at 0.1, 0.2).}
\label{fig:colored_mnist}
\vspace{-1.5ex}
\end{figure*}

\begin{figure*}[t!]
\centering
\includegraphics[width=0.77\textwidth]{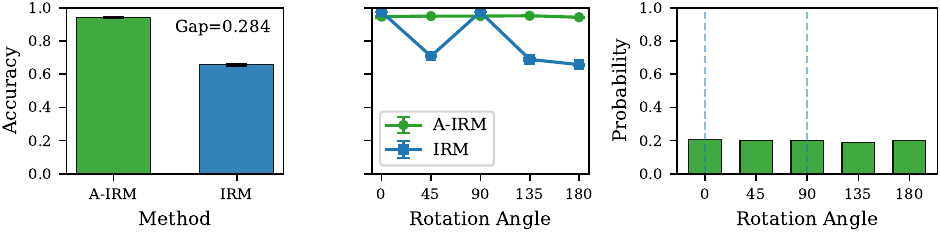}
\vspace{-1.5ex}
\caption{\small \airm environment discovery on \rotatedmnist. \textbf{(Left)} Worst-case accuracy comparison shows 28.4\% improvement (94.2\% vs 65.8\%). \textbf{(Middle)} Test accuracy across rotation angles—\irm fails catastrophically on unseen rotations (45°, 135°, 180°) while \airm maintains consistent performance. \textbf{(Right)} \airm learns approximately uniform distribution over all rotations, unlike \irm which only uses 0° and 90° (indicated by dashed lines).}
\label{fig:rotated_mnist}
\vspace{-1.5ex}
\end{figure*}

Next, we analyze the environment policies learned by \airm on the synthetic benchmarks to gain insight into its superior performance.
The key finding is that \airm successfully identifies the most informative environment distributions for robust learning, which often differ substantially from those used in standard, fixed-\irm settings. 
\cref{fig:colored_mnist} and \cref{fig:rotated_mnist} show the analyses on \coloredmnist and \rotatedmnist, respectively.
We observe that, on \coloredmnist, \airm discovers that intermediate correlations ($e \approx 0.35$) provide the optimal learning signal—avoiding both uninformative similar environments and conflicting diverse environments.
Meanwhile, on \rotatedmnist, \airm learns approximately uniform distribution over all rotations, leading to robust performance across all test angles while IRM catastrophically fails on unseen rotations.

\subsection{Theoretical Validation: Variance Regularization}

\begin{figure}[h]
    \centering
    \begin{subfigure}[c]{0.232\textwidth}
        \centering
        \includegraphics[width=0.99\textwidth]{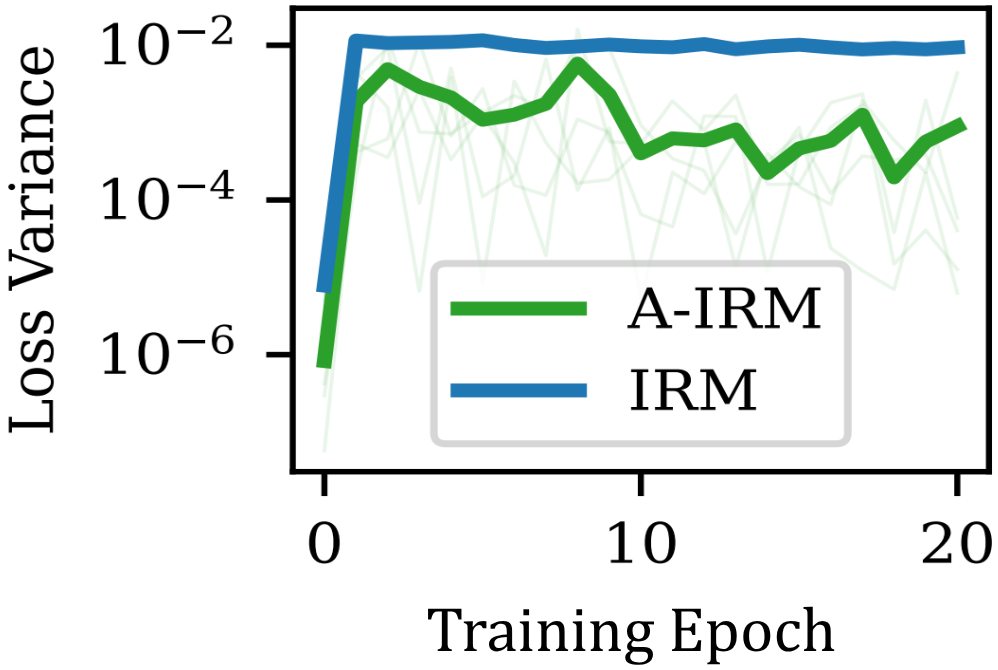}
    \end{subfigure}
    \hfill
    \begin{subfigure}[c]{0.23\textwidth}
        \centering
        \includegraphics[width=0.99\textwidth]{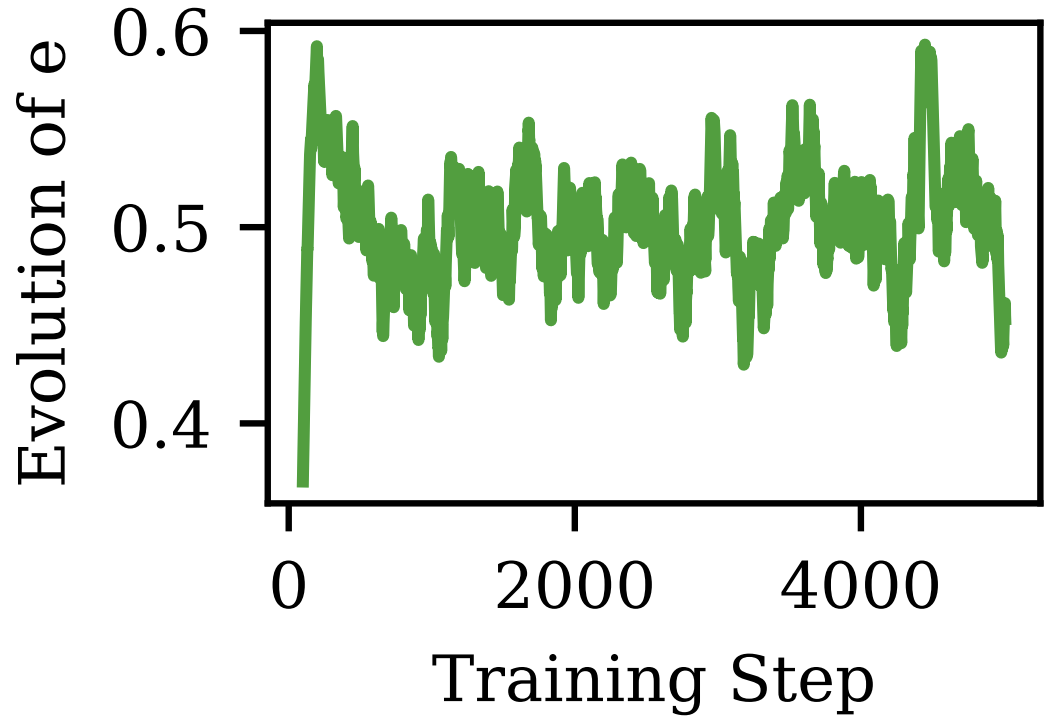}
    \end{subfigure}
    \vspace{-1.5ex}
    \caption{Evolution of the loss variance of \airm and \irm (\textbf{Left}), and the environment $e$ of \airm (\textbf{Right}) on \coloredmnist.}
    \label{fig:evolition_colored_mnist}
    \vspace{-2.5ex}
\end{figure}




Our theoretical analysis (\cref{prop:loa-la-main}) predicts that \airm's loss-averaging objective implicitly minimizes cross-environment variance. \cref{fig:evolition_colored_mnist}-(left) provides striking empirical validation: \airm achieves approximately 100× lower final loss variance ($\sim 10^{-4}$) compared to \irm ($\sim 10^{-2}$). This dramatic variance reduction explains why \airm discovers solutions that perform consistently across all environments rather than overfitting to specific ones.

\subsection{Results on Waterbirds Dataset}

\begin{table}[h]
\centering
\caption{\small{Performance on \waterbirds dataset. All adaptive variants improve upon their baselines.}}
\label{tab:waterbirds}
\vspace{-1.5ex}
\renewcommand{\arraystretch}{1.00}
\scalebox{.85}{
\begin{tabular}{llcccc}
\toprule
\multirow{2}{*}{\textbf{Method}} & \multirow{2}{*}{\textbf{Type}} & \multicolumn{2}{c}{\textbf{ResNet18}} & \multicolumn{2}{c}{\textbf{ResNet50}} \\
\cmidrule(lr){3-4} \cmidrule(lr){5-6}
& & Mean & Worst & Mean & Worst \\
\midrule
\midrule
\erm & Baseline & 85.2 & 60.0 & 87.1 & 63.2 \\
\midrule
\groupdro & Baseline & 88.0 & 78.0 & 93.2$^*$ & 86.0$^*$ \\
\agroupdro & Adaptive & \textbf{88.8} & \textbf{78.0} & \textbf{92.1} & \textbf{87.8} \\
\midrule
\rex/VREx & Baseline & 87.0 & 74.0 & — & — \\
\arex & Adaptive & \textbf{88.8} & \textbf{78.0} & \textbf{92.7} & 80.5 \\
\midrule
\coral & Baseline & 86.0 & 73.0 & — & — \\
\acoral & Adaptive & \textbf{86.0} & \textbf{75.0} & \textbf{90.8} & 82.9 \\
\bottomrule
\end{tabular}
}
\vspace{2pt}
{\footnotesize $^*$\groupdro baseline from \citet{sagawa2019distributionally}.}
\vspace{-2ex}
\end{table}

Results on \waterbirds dataset (\cref{tab:waterbirds}) demonstrate that our \uaed framework provides a consistent performance boost across multiple robust learning methods. In all tested scenarios (\erm, \groupdro, \rex, \coral) and architectures (\resneteighteen, \resnetfifty), the Adaptive (A-) variants match or significantly exceed the worst-group accuracy of their vanilla baselines.
Notably, \agroupdro achieves the highest worst-case accuracy of $87.8\%$ (\resnetfifty), surpassing the strong \groupdro baseline (86.0\%) and confirming the efficacy of adaptively learning environment distributions—even when known groups exist.
This universal improvement shows that \uaed functions as a generic enhancement for targeting and improving generalization to the most vulnerable data subgroups. For additional experiments on PACS dataset \citep{li2017deeperbroaderartierdomain}, see \cref{add_experiments}.


\subsection{Comparison with State-of-the-Art}

\begin{table}[h]
\centering
\caption{\small{Comparison with SOTA methods on Waterbirds (worst-group accuracy) using \resnetfifty.
$^\dagger$Uses additional concept discovery. $^\ddagger$Requires retraining last layer.
}
}
\label{tab:sota}
\vspace{-1.5ex}
\renewcommand{\arraystretch}{1.00}
\scalebox{.91}{
\begin{tabular}{lc}
\toprule
\textbf{Method}  & \textbf{Worst-Group Acc} \\
\midrule
\midrule
\groupdro \citep{sagawa2019distributionally}  & 86.0 \% \\
\jtt \citep{liu2021just}  & 86.7 \% \\
\disc \citep{wu2023discovering} & 88.0 \% $^\dagger$ \\
\dfr \citep{kirichenko2023last} & 91.2\% $^\ddagger$ \\
\midrule
\agroupdro \textbf{(Ours)} & 87.8 \% \\
\bottomrule
\end{tabular}
}
\vspace{-1.5ex}
\end{table}


To contextualize the performance of our \uaed framework, \cref{tab:sota} compares the worst-group accuracy of our top performing variant, \agroupdro, against \gls{SOTA} methods on the \waterbirds dataset using \resnetfifty.
Our \agroupdro variant achieves a highly competitive worst-group accuracy of $87.8\%$, positioning the \uaed framework favorably among specialized \gls{SOTA} methods.
Crucially, while methods like \groupdro, \jtt, \disc, and \dfr either rely on external group annotations, require complex two-stage training procedures, or introduce architecture modifications,
\agroupdro achieves result using a simple, single-stage, alternating optimization over parameterized transformations.
This competitive performance, achieved without the need of pre-defined groups or significant procedure overhead, highlights the advantage of environment adaptivity as a generic, assumption-free enhancement to robust learning.

\begin{table}[h]
\centering
\caption{\small Ablation study on \airm (\coloredmnist).}
\label{tab:ablation_airm}
\vspace{-1.5ex}
\renewcommand{\arraystretch}{1.00}
\scalebox{.89}{
    \begin{tabular}{lcc}
    \toprule
\multicolumn{1}{c}{\textbf{Method}} & \multicolumn{1}{c}{\begin{tabular}[c]{@{}c@{}}\textbf{Worst-Case}\\ \textbf{Acc} (\%)\end{tabular}} & \multicolumn{1}{c}{\begin{tabular}[c]{@{}c@{}}\textbf{Mean} $e$\\ \textbf{Discovered}\end{tabular}} \\
    \midrule
    \midrule
    \airm (full) & $72.3 \pm 1.6$ & $0.354 \pm 0.023$ \\
    \quad w/o hierarchical policy & $68.9 \pm 2.1$ & $0.287 \pm 0.041$ \\
    \quad w/o KL regularization & $67.5 \pm 2.8$ & $0.198 \pm 0.018$ \\
    \quad w/ fixed variance & $69.5 \pm 1.9$ & $0.312 \pm 0.015$ \\
    \midrule
    Different MC samples: & & \\
    \quad K=1 & $70.1 \pm 2.0$ & $0.341 \pm 0.029$ \\
    \quad K=5 (default) & $71.4 \pm 1.6$ & $0.354 \pm 0.023$ \\
    \quad K=10 & $71.6 \pm 1.5$ & $0.356 \pm 0.022$ \\
    \bottomrule
    \end{tabular}
}
\vspace{-1ex}
\end{table}

\begin{table*}[h]
\centering
\caption{\small Discovered environment statistics across methods and datasets}
\label{tab:env_discovery}
\vspace{-1.5ex}
\renewcommand{\arraystretch}{1.00}
\scalebox{.92}{
\begin{tabular}{llp{12cm}}
\toprule
\textbf{Method} & \textbf{Dataset} & \textbf{Discovered Pattern} \\
\midrule
\airm & \coloredmnist & Concentrates on $e \approx 0.35$: intermediate correlations that balance breaking spurious patterns with maintaining learnability \\
\airm & \rotatedmnist & Uniform distribution over all rotations: maximizes diversity for invariance \\
\midrule
\agroupdro & \waterbirds & Focuses on minority groups with high spurious correlation \\
\arex & \waterbirds & Discovers maximally diverse environments while maintaining stability \\
\acoral & \waterbirds & Identifies environments with distinct feature distributions \\
\bottomrule
\end{tabular}
}
\vspace{-1.5ex}
\end{table*}

\subsection{Ablation Studies}

\begin{table}[h]
\centering
\caption{\small{Ablation study on adaptive methods for \waterbirds (\resneteighteen).}}
\label{tab:ablation_waterbirds}
\vspace{-1.5ex}
\renewcommand{\arraystretch}{1.00}
\scalebox{.77}{
\begin{tabular}{llcc}
\toprule
\multicolumn{1}{c}{\textbf{Method}} & \multicolumn{1}{c}{\textbf{Configuration}} &
\multicolumn{1}{c}{\begin{tabular}[c]{@{}c@{}}\textbf{Mean}\\ \textbf{Acc} (\%)\end{tabular}}
&\multicolumn{1}{c}{\begin{tabular}[c]{@{}c@{}}\textbf{Worst-group}\\ \textbf{Acc} (\%)\end{tabular}} \\
\midrule
\midrule
\multirow{3}{*}{\agroupdro
} 
& Full model & 88.8 & 78.0 \\
& w/o adaptive policy & 88.0 & 78.0 \\
& w/o KL regularization & 87.5 & 75.2 \\
\midrule
\multirow{3}{*}{\arex} 
& Full model & 88.8 & 78.0 \\
& w/o variance penalty & 87.2 & 74.5 \\
& Fixed environments & 87.0 & 74.0 \\
\midrule
\multirow{3}{*}{\acoral}
& Full model & 86.0 & 75.0 \\
& w/o MMD alignment & 85.5 & 73.2 \\
& Fixed environments & 86.0 & 73.0 \\
\bottomrule
\end{tabular}
}
\vspace{-2.5ex}
\end{table}


Ablation studies across \coloredmnist (\cref{tab:ablation_airm}) and \waterbirds (\cref{tab:ablation_waterbirds}) confirm the necessity of the core components of our \uaed framework.
In particular, the hierarchical policy structure is crucial, as its removal leads to a $3.4\%$ drop in \airm's worst-case accuracy, underscoring its role in managing the large environment space.
Furthermore, the KL regularization proves essential, preventing the environment policy $\Pi_{\phi}$ from collapsing onto overly simple or challenging single environments, evidenced by a significant drop ($4.8\%$ for \airm) and a lower mean discovered environment factor ($0.354 \rightarrow 0.198$).
In addition, the number of MC samples, $K$, shows diminishing returns, with $K=5$ striking an optimal balance.
Finally, the analysis of discovered environments (\cref{tab:ablation_airm} reveals that \uaed's policy is not random but interpretable and objective-specific: \airm on \coloredmnist concentrates on an intermediate correlation ($e \approx 0.35$) that maximizes the \irm penalty, while \agroupdro on \waterbirds focuses its probability mass on the known minor groups, effectively reproducing the ideal \groupdro behavior without explicit group labels.
These results validate our design choices and further confirm that the adaptive policy successfully discovers robust and meaningful environmental variations.

\subsection{Detailed Environment Discovery Analysis}

The detailed analysis of the learned environment policies (\cref{tab:env_discovery}) provides crucial, interpretable evidence for the efficacy of our \uaed framework.
The resulting environment distributions are not generic but are precisely tailored to the objective of the base robust method. 
This validation confirms that our framework successfully finds the ideal challenging distribution $\Pi_{\phi}$ that maximizes the specific robustness criteria for each baseline. This is evidenced by the policy's ability to find the maximal diversity required for general invariance tasks (like \rotatedmnist) or to learn to focus on the most critical data subgroups (like \waterbirds), effectively automating complex environment specification without relying on external annotations.


\subsection{Discussion: Why Universal Adaptive Discovery Works}

Our experiments highlight three factors behind the effectiveness of adaptive environment discovery.  
\emph{(1) Method–Environment Alignment:} Different robust objectives benefit from different environments—e.g., \agroupdro discovers worst-case scenarios, while \arex favors diverse yet learnable ones. The adaptive framework lets each method find the environments best suited to its objective.  
\emph{(2) Continuous Exploration:} Unlike fixed specifications, adaptive discovery updates the environment distribution in response to model weaknesses, creating a curriculum that progressively challenges the learner.  
\emph{(3) Implicit Regularization:} The hierarchical Bayesian prior and KL regularization discourage overfitting to particular environments while promoting exploration of diverse conditions.









\vspace{-1ex}

\section{Conclusion}

\vspace{-1ex}

We introduced \emph{Universal Adaptive Environment Discovery} (\uaed), a unified framework that shifts robust learning from relying on fixed environments to discovering them adaptively. Our work shows that diverse robust methods such as \irm, \rex, \groupdro, and \coral can all benefit from adaptive environment discovery, supported by theoretical guarantees of distributionally robust generalization and empirical validation on synthetic and real-world data. Beyond empirical and theoretical contributions, we provide the conceptual insight that environment specification and robust objectives are complementary aspects of the same problem. These results suggest that the future of robust learning lies not in designing new objectives or manually defining environments, but in methods that jointly discover both—making environment discovery adaptive and automatic to provide a principled path toward truly robust machine learning systems.

\paragraph{Limitations and future work.}
Adaptive methods incur higher computational cost by sampling $K$ environments per batch, though this overhead can be mitigated via parallelization. While environments are learned, the transformation family $T_\gamma$ must still be specified, suggesting future work on learning transformation functions directly. Finally, our theory guarantees robustness within KL-balls of the learned distribution, but connecting these guarantees to worst-case out-of-distribution (OOD) performance remains an open question.







\newpage

\bibliography{refs}

\clearpage
\appendix
\thispagestyle{empty}

\onecolumn
\aistatstitle{Universal Adaptive Environment Discovery: \\
Supplementary Materials}

\section{Theoretical Proofs}
\label{app:theory_proof}
\subsection{Preliminaries}
We use the following standard facts.

\begin{lemma}[Hoeffding's lemma]\label{lem:hoeffding}
If $X\in[a,b]$, then $\log \E e^{\lambda(X-\E X)} \le \frac{\lambda^2(b-a)^2}{8}$ for all $\lambda\in\R$.
\end{lemma}

\begin{lemma}[Donsker--Varadhan (DV)]\label{lem:dv}
For probability measures $G,\Pi$ and any measurable $g$,
\(
\E_G[g]
\le \frac{ \KL(G\Vert \Pi) + \log \E_\Pi e^{\lambda g} }{\lambda},\ \forall\lambda>0.
\)
\end{lemma}

\subsection{Proof of Theorem~\ref{thm:pacbayes-policy-main}}
Consider the composite bounded loss
\(
\tilde\ell(h,\phi;z)=\E_{\gamma\sim\Pi_\phi}\E_{z'|z}\ell(h,z')\in[0,1].
\)
Apply the classical PAC--Bayes bound (e.g., for bounded losses) to the joint hypothesis $H=(h,\phi)$ with prior $M=P\times\Pi_0$ and posterior $Q$:
with probability $\ge 1-\delta$,
\[
\E_{H\sim Q}\E_{z\sim P}\tilde\ell(H;z)
\le
\E_{H\sim Q}\frac1n\sum_{i=1}^n \tilde\ell(H;z_i)
+\sqrt{\frac{\KL(Q\Vert M)+\ln(1/\delta)}{2n}}.
\]
Identifying the terms with $\cR_{\Pi_\phi}$ and $\hat\cR_{\Pi_\phi}$ completes the proof. \qed

\subsection{Proof of Lemma~\ref{lem:transport-main}}
Let $f\in[0,1]$. By Lemma~\ref{lem:dv}, for any $\lambda>0$,
\(
\E_G f \le \frac{\KL(G\Vert\Pi_\phi)+\log \E_{\Pi_\phi}e^{\lambda f}}{\lambda}.
\)
Write $f=(f-\E_{\Pi_\phi}f)+\E_{\Pi_\phi}f$ and apply Lemma~\ref{lem:hoeffding} with $a=0,b=1$:
\(
\log \E_{\Pi_\phi}e^{\lambda(f-\E f)} \le \lambda^2/8.
\)
Hence
\(
\E_G f \le \E_{\Pi_\phi} f + \frac{\KL(G\Vert\Pi_\phi)}{\lambda} + \frac{\lambda}{8}.
\)
Optimizing over $\lambda>0$ gives $\lambda^\star=\sqrt{8\,\KL(G\Vert\Pi_\phi)}$ and the value
$\E_{\Pi_\phi} f + \sqrt{\tfrac12\,\KL(G\Vert\Pi_\phi)}$. \qed

\subsection{Proof of Theorem~\ref{thm:universal-robust-main}}
Fix $Q$. Let $f_H(\gamma)=\cR^{e(\gamma)}(h)$ for $H=(h,\phi)$. By Lemma~\ref{lem:transport-main},
\[
\E_{H\sim Q}\E_{\gamma\sim G} f_H(\gamma)
\le
\E_{H\sim Q}\E_{\gamma\sim \Pi_\phi} f_H(\gamma)
+ \sqrt{\tfrac12\,\KL(G\Vert \Pi_\phi)}.
\]
Apply Theorem~\ref{thm:pacbayes-policy-main} to the first term and use $\KL(G\Vert\Pi_\phi)\le\rho$. \qed

\subsection{Proof of Proposition~\ref{prop:entropic-duality-main}}
By DV (Lemma~\ref{lem:dv}),
\[
\sup_{G:\,\KL(G\Vert\Pi_\phi)\le\rho}\E_G r_\gamma
=
\inf_{\lambda>0}\sup_G
\frac{\KL(G\Vert\Pi_\phi)-\rho + \log \E_{\Pi_\phi}e^{\lambda r_\gamma}}{\lambda}
=
\inf_{\lambda>0}\frac{\rho + \log \E_{\Pi_\phi}e^{\lambda r_\gamma}}{\lambda},
\]
where we used the Fenchel dual of the indicator of the KL--ball. For the finite-support bound, note that for uniform $\Upsilon$ on the support ($|\supp|=k$),
\[
\max_\gamma r_\gamma
\le \frac{1}{\lambda}\log \sum_{\gamma}\Upsilon(\gamma)\,e^{\lambda r_\gamma}
+ \frac{\KL(\Upsilon\Vert \text{Unif})}{\lambda}
= \frac{1}{\lambda}\log\Big(\tfrac1k \sum_{\gamma}e^{\lambda r_\gamma}\Big) + \frac{\log k}{\lambda}.
\]
\qed

\subsection{Proof of Proposition~\ref{prop:loa-la-main}}
By Jensen, $\cL_{\text{LA}}\le \cL_{\text{LoA}}$, hence the gap is nonnegative. Let $Y=\ell_\gamma-\mu$; then
\(
\cL_{\text{LoA}}-\cL_{\text{LA}}
= \log \E e^{-Y}.
\)
If $Y$ is sub-Gaussian with proxy $\sigma^2$, then $\log \E e^{tY}\le \sigma^2 t^2/2$ for all $t\in\R$. With $t=-1$ this gives the stated upper bound. \qed


\subsection{Convergence Analysis of the Alternating Optimization Strategy in \cref{sec:optim_strategy}} \label{app:alternating_optim}

Our joint objective function of the model parameters $\theta$ (for hypothesis $h_{\theta}$ and the policy parameters $\phi$ (for environment distribution $\Pi_{\phi}$) is defined follows:
\begin{align}
L(\theta,\phi) =  \E_{\gamma\sim \Pi_\phi}\E_{z\sim P}\ell(h_\theta,T_\gamma(z)) + \mathcal{P}(h_\theta, \Phi_\phi) + \beta \cdot \,\KL(\Pi_\phi\Vert \Pi_0). \nonumber
\end{align}
\begin{assumption}[Smoothness and gradient noise]\label{ass:sgd-main}
$L$ is lower bounded and has $L$-Lipschitz gradient; the reparameterized-gradient estimators for $(\theta,\phi)$ are unbiased with bounded variance; step sizes satisfy $\sum_t\eta_t=\infty$, $\sum_t\eta_t^2<\infty$.
\end{assumption}

\begin{theorem}[Nonconvex alternating SGD]\label{thm:convergence-main}
Under Assumption~\ref{ass:sgd-main}, alternating stochastic gradient updates on $(\theta,\phi)$ satisfy
\[
\frac1T\sum_{t=1}^T \E\big[\|\nabla L(\theta_t,\phi_t)\|^2\big] \;=\; \cO(T^{-1/2}).
\]
\end{theorem}

\textit{Proof.}
Under Assumption~\ref{ass:sgd-main}, the standard descent lemma for $L$ with $L$-Lipschitz gradient yields, for the alternating update,
\[
\E[L_{t+1}] \le \E[L_t] - \tfrac{\eta_t}{2}\E\|\nabla L_t\|^2 + C\eta_t^2,
\]
for some constant $C$ depending on the gradient-noise variance. Summing and using $\sum_t\eta_t=\infty$, $\sum_t\eta_t^2<\infty$ gives
\(
\frac{1}{\sum_{t=1}^T \eta_t}\sum_{t=1}^T \eta_t\,\E\|\nabla L_t\|^2 \le \cO(1/\sum_{t=1}^T \eta_t)+\cO(\tfrac{\sum_{t=1}^T \eta_t^2}{\sum_{t=1}^T \eta_t}).
\)
With $\eta_t=\eta/\sqrt{t}$ this becomes $\cO(T^{-1/2})$. \qed

\section{Extended Experimental Details}
\label{app:experiment_details}
\subsection{Implementation Details}

\textbf{Architecture Details:}
\begin{itemize}
\item \resneteighteen/\resnetfifty: Pretrained on \imagenet, final layer replaced for binary classification
\item Feature dimension: 512 for \resneteighteen, 2048 for \resnetfifty
\item Dropout: 0.0 (following \groupdro setup)
\end{itemize}

\textbf{Training Details:}
\begin{itemize}
\item Optimizer: Adam with $\beta_1=0.9$, $\beta_2=0.999$
\item Learning rate: $10^{-4}$ for model, $10^{-3}$ for policy parameters
\item Weight decay: $10^{-5}$
\item Batch size: 128
\item Number of epochs: 30
\item Early stopping: Based on worst-group validation accuracy
\end{itemize}

\textbf{Adaptive Policy Configuration:}
\begin{itemize}
\item Continuous policy: Beta distribution with learnable $\alpha, \beta$ parameters
\item Policy network: 2-layer MLP with 64 hidden units
\item KL regularization weight $\alpha$: 1.0
\item Monte Carlo samples: 3 per batch
\item Warm-up: 5 epochs before enabling adaptive policy
\end{itemize}

\subsection{Dataset Details}

\textbf{Waterbirds:}
\begin{itemize}
\item Training: 4,795 samples
\item Validation: 1,199 samples  
\item Test: 5,794 samples
\item Groups: 4 groups based on (bird type, background) combinations
\item Group distribution: Highly imbalanced with smallest group having $<$ 100 training samples
\end{itemize}

\paragraph{Assets \& Licenses.}
We use only publicly available research datasets and pretrained models:

\begin{itemize}
\item \textbf{CUB-200-2011} \citep{wah2011caltech}: license published by creators (non-commercial academic use).
\item \textbf{Waterbirds} \citep{sagawa2020waterbirds}: derived from CUB and Places; follows the respective research-use terms.
\item \textbf{MNIST} (Colored/Rotated variants): generated from MNIST via our scripts; inherits MNIST’s research-use terms.
\item \textbf{Pretrained ResNet-18/50} (PyTorch/torchvision): model weights and code under the PyTorch/torchvision license.
\end{itemize}

We release only \emph{code/configs} to reproduce results (no new datasets or human/PII data).
No additional consent was sought or required beyond the datasets’ published licenses/terms.
The experiments contain no personally identifiable or offensive content.




\section{Additional Experimental Results}
\label{add_experiments}
\subsection{Results on PACS Dataset \citep{li2017deeperbroaderartierdomain}}

\begin{table}[h]
\centering
\caption{Results on PACS dataset}
\label{tab:pacs}
\begin{tabular}{lccccc}
\toprule
\textbf{Method} & \textbf{Photo} & \textbf{Art} & \textbf{Cartoon} & \textbf{Sketch} & \textbf{Average} \\
\midrule
\agroupdro & 85.84 & 77.09 & 97.66 & 80.40 & 85.25 \\
\arex & 85.45 & 74.06 & 97.49 & 79.10 & 84.16 \\
\acoral & 84.80 & 73.95 & 97.11 & 79.05 & 83.73 \\
\bottomrule
\end{tabular}
\end{table}

\subsection{Additional Ablation Studies}

\begin{table}[H]
\centering
\caption{Effect of different transformation families on \waterbirds}
\begin{tabular}{lcc}
\toprule
\textbf{Transformation Type} & \textbf{Mean Acc} & \textbf{Worst-Group Acc} \\
\midrule
Correlation strength (continuous) & 88.8 & 78.0 \\
Discrete groups & 87.9 & 76.5 \\
\bottomrule
\end{tabular}
\end{table}

\section{Extended Related Work}

\subsection{Connection to Meta-Learning}

Our \uaed framework shares conceptual similarities with the bi-level optimization common in meta-learning approaches \citep{finn2017model, nichol2018reptile}, as both involve learning a higher-level policy ($\phi$) to guide a lower-level model ($\theta$). However, their objectives differ: Meta-learning seeks adaptability—finding a strategy (e.g., an initialization) for fast learning on new tasks with few examples—while \uaed seeks invariance by discovering the optimal distribution of synthetic environments necessary to enforce a robust predictor that generalizes across all potential distribution shifts.

\subsection{Connection to Curriculum Learning}

Our adaptive framework \uaed implicitly implements a form of curriculum learning \citep{bengio2009curriculum} by dynamically adjusting the environment policy $\Pi_{\phi}$ throughout training. In the early training phase, the policy explores diverse environments, easing the initial learning task. During mid training, the focus shifts to more challenging yet informative and learnable environments, ensuring the model efficiently extracts the invariant signal. Finally, in the late training phase, the policy concentrates on the worst-case scenarios (e.g., maximizing risk variance for \arex or the max risk for \agroupdro), effectively stress-testing the mature model to achieve maximal distributional robustness and pushing the invariant predictor to its generalization limits. This continuous, self-paced adjustment of the training distribution is key to the framework's stability and superior performance.

\subsection{Domain Adaptation vs Environment Discovery}

Our \uaed framework differs fundamentally from traditional \gls{DA} \citep{ganin2016domain}. While \gls{DA} relies on access to a specific target domain's data to align features, \uaed requires no target domain; instead, it actively discovers and generates multiple synthetic environments via parameterized kernels. Crucially, \uaed optimizes for worst-case robustness across the learned environment distribution, aiming for general invariance, rather than targeting performance on a single, fixed distribution.

\end{document}